*Article*

# Quality over Quantity: An Effective Large-Scale Data Reduction Strategy Based on Pointwise V-Information

**Fei Chen * and Wenchi Zhou**

Information Science and Technology College, Dalian Maritime University, Dalian 116026, China; zhouwc@dlmu.edu.cn
* Correspondence: chenf@dlmu.edu.cn

**Abstract**

In order to increase the effectiveness of model training, data reduction is essential to data-centric Artificial Intelligence (AI). It achieves this by locating the most instructive examples in massive datasets. To increase data quality and training efficiency, the main difficulty is choosing the best examples rather than the complete datasets. In this paper, we propose an effective data reduction strategy based on Pointwise $\mathcal{V}$-Information (PVI). To enable a static method, we first use PVI to quantify instance difficulty and remove instances with low difficulty. Experiments show that classifier performance is maintained with only a 0.0001% to 0.76% decline in accuracy when 10%–30% of the data is removed. Second, we train the classifiers using a progressive learning strategy on examples sorted by increasing PVI, accelerating convergence and achieving a 0.8% accuracy gain over conventional training. Our findings imply that training a classifier on the chosen optimal subset may improve model performance and increase training efficiency when combined with an efficient data reduction strategy. Furthermore, we have adapted the PVI framework, which was previously limited to English datasets, to a variety of Chinese Natural Language Processing (NLP) tasks and base models, yielding insightful results for faster training and cross-lingual data reduction.

## 1. Introduction

Driven by large-scale datasets, large language models, pre-training, and finetuning training procedures, Artificial Intelligence (AI) technology has made remarkable progress in the field of Natural Language Processing (NLP). With the widespread application of AI systems, it has become increasingly clear that model performance is greatly influenced by the quality of data [1]. Traditionally, the Model-Centric paradigm has been used to conduct AI research, with a primary focus on developing new model architectures and suggesting optimal algorithms to boost performance [2]. However, this paradigm often overlooks the intrinsic quality of data. Issues including imbalance, labeling errors, and redundant data may result in deteriorated model performance, skewed outcomes, and poor decision-making [3–5]. As they say, "garbage in, garbage out." It is more beneficial to improve the quality of data than to merely increase its quantity [6,7]. As a result, a new paradigm known as data-centric AI has surfaced, emphasizing systematic improvement in data quality to enhance model performance using fewer but high-quality data. The importance of the data-centric paradigm is being recognized, advocating a shift in focus from continuously improving model architectures and algorithm optimization to prioritizing the enhancement of high-quality data [8].

Data reduction is a fundamental strategy in data-centric AI [9], which assesses data quality, eliminates low-quality data, and retains predominantly high-quality data in a dataset to optimize model training efficacy and performance. One of the representative methods for dataset reduction is dataset distillation, which aims to improve data processing efficiency by synthesizing a small typical dataset from substantial data [10]. The training of large language models often relies on large-scale datasets [11], which may diminish training efficiency and model generalization due to the presence of redundant low-quality data and incur significant computational and storage costs. To reduce the training dataset to a manageable size while maintaining model performance, the primary challenge in data reduction is selecting the optimal subset from large-scale datasets utilizing data quality measures as a reference.

The term dataset difficulty pertains to the quality of the data. It is fundamentally rooted in information theory and serves in the generalization of mutual information. It assesses the informational richness or learning complexity of data. The assumption of dataset difficulty is that complex knowledge and challenging data contribute to the development of powerful models whereas an abundance of low-quality data impedes the model learning efficiency.

Several measures have been proposed as potential indicators of a dataset's difficulty. Devin Kwok [12] focused on example difficulty scores, such as Prediction Depth [13] and Variance of Gradients (VoG) [14]. Peng Cui et al. [15] evaluated sample difficulty by employing feature-space Gaussian modeling and relative Martens distance calculation. David Mayo et al. [16] introduced Minimum Viewing Time as a dataset difficulty measure. Chengwen Wang et al. [17] proposed four difficulty measures to be applied to named entity recognition datasets, including three internal measures (invisible entity ratio, entity ambiguity, and text complexity) and one external measure (model variance).

Pointwise $\mathcal{V}$-Information (PVI) [18] is a promising metric for quantifying dataset difficulty which defines dataset difficulty as the lack of model-usable information. After processing the original data $X$, the increase in the ability of humans or machine learning algorithms to predict the label $Y$ is the usable information. More formally, let $t$ be the decryption algorithm and $\mathcal{V}$ be a class of processing functions. It follows that $I_{\mathcal{V}}(t(X) \rightarrow Y) > I_{\mathcal{V}}(X \rightarrow Y) \approx 0$. The less usable information there is, the more difficult the dataset is for a model. Furthermore, PVI could measure the difficulty of each instance in a given dataset. Within this framework, instance difficulty is defined as the inherent challenge that an individual data point presents to a specific model in learning or prediction. The instance difficulty is inversely proportional to the amount of $\mathcal{V}$-usable Information the instance provides. There is a difference between instance difficulty and instance usefulness; the latter should be the contribution an individual data instance makes to a model's generalization capability and overall performance. An instance's usefulness is a multifaceted concept that depends on both its inherent difficulty and its relevance to the specific learning task. For data instances, high PVI indicates that they are easy for the model to learn. During training, a small number of easy instances can elevate model performance to a certain level, but continuously feeding easy instances yields minimal performance gains. Low PVI indicates that the instances are hard to learn but could gain a larger margin of performance than the easy instances. This suggests that excessive easy instances, marked by high PVI, are redundant, leading to a significant waste of computational resources and disproportional performance gains.

The PVI framework provides new perspectives on dataset evaluation, selection, and reduction. However, since most research has been conducted on English datasets, there were concerns about its applications in a cross-lingual context. Is it possible to generalize the data reduction approach and dataset difficulty measures to other languages? How can we use PVI to boost model performance in a cross-lingual context and increase training efficiency?

To address these issues, we present a PVI-based large-scale data reduction strategy. Our research focuses on obtaining an optimal subset for training while improving training efficiencies, saving computational resources, and preserving model performance. The contributions of our paper are as follows:

- We eliminated low-quality instances by employing PVI to assess instance difficulty. Experiments demonstrate that the removal of 10%–30% of the data results in a minimal decline in classifier performance, ranging from 0.0001% to 0.76% in accuracy. This indicates that we might effectively sustain model performance and accelerate training by removing a certain quantity of low-quality instances.

- We proposed a progressive learning technique that trained a classifier by initially organizing examples in descending order of PVI. This challenging training method not only accelerated model convergence but also resulted in a 0.8% enhancement in accuracy. The results indicate that strategically employing PVI to guide the training process could significantly enhance model performance and training efficiency. In contrast to current curriculum learning methods that depend on heuristic or less universal difficulty metrics, our approach utilizes the robust and adaptable characteristics of PVI to effectively direct the learning process, resulting in enhanced training efficiency and superior generalization for large language models.

- The PVI framework, hitherto restricted to English datasets, has been adapted for various Chinese NLP tasks and fundamental models. The cross-lingual extension provides novel perspectives for data-centric AI in broader application contexts, confirms the universality of the PVI framework, and supplies valuable insights for cross-lingual data reduction.

## 2. Materials and Methods

Alex Havrilla [19] notes, in his definition of dataset complexity, that complexity is a data characteristic that intuitively reflects the difficulty of a sample. He defines the complexity $C(\omega)$ of an instance $\omega \in \Omega$ as its size under a fixed representation scheme. An n-sample complexity measure is represented as a function $C: \Omega^n \to \mathbb{R}$, which intuitively measures the difficulty of the data, defining complexity $C_\Omega \to \mathbb{R}$ at the level of a single sample, with $C$ being recovered as the average over samples.

Several fixed representation schemes mentioned above exhibit certain limitations. For instance, the example difficulty scores used by Devin Kwok include various scores for quantifying the difficulty of individual instances in the training dataset, which typically depend on the model. The relative Martens distance calculation used by Peng Cui et al. is primarily applied in computer vision tasks such as image classification. The Minimum Viewing Time introduced by David Mayo et al. is also limited to quantifying the difficulty of computer vision datasets [20,21]. In NLP, the entity ratio, entity ambiguity, text complexity, and model variance used by Chengwen Wang et al. are only targeted at named entity recognition datasets [22,23].

In contrast, PVI offers a more universal and flexible approach to measuring the instance difficulty. PVI quantifies the difficulty of the individual instance within a given distribution, framing dataset difficulty with respect to a model $\mathcal{V}$. Dataset difficulty is conceptualized as the lack of information readily usable by model $\mathcal{V}$. A significant advantage of PVI is its ability to facilitate cross-dataset difficulty comparisons, even across diverse label spaces. This inherent flexibility provides PVI with a much broader application scope compared to the traditional performance metrics.

The escalating scale of modern datasets poses substantial challenges for model training, which necessitates immense computational resources and prolonged training times. While existing methods offer valuable insights into dataset difficulty, their inherent limitations often restrict their applicability to diverse and large-scale scenarios. PVI addresses these challenges by providing a model-aware quantification of individual instance difficulty. By identifying and prioritizing data instances based on their PVI, we can significantly streamline the training process, and enable data reduction while maintaining model performance.

## 2.1. Model Architecture

The purpose of this paper is to construct an efficient data reduction strategy to optimize the efficiency of data usage in natural language inference (NLI) [24] tasks by quantifying and using the difficulty of data instances. To this end, we designed a comprehensive framework that includes three modules: **data transformation**, **PVI calculator**, and **reduction approach**. The overall architecture is shown in Figure 1.

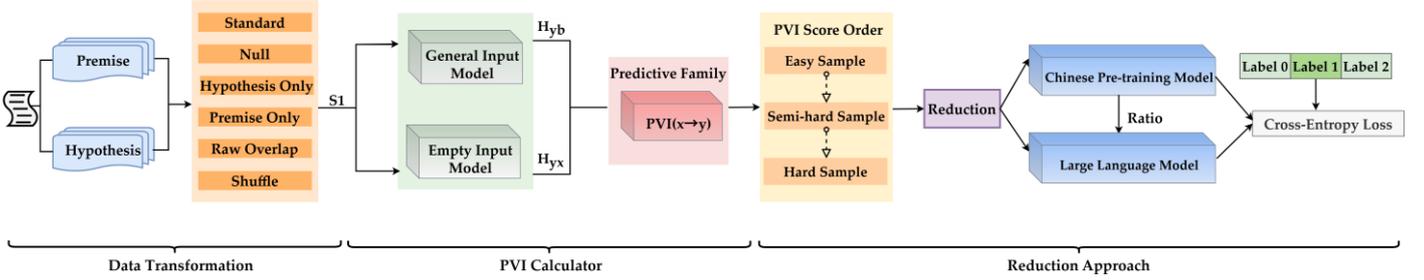

**Figure 1.** The model architecture of the data reduction strategy.

**Data Transformation:** This module is the preprocessing stage of the entire process and is responsible for converting the original dataset into a variety of input formats required by subsequent modules. It is an NLI Transformation base class, which defines standard processes for data loading, filtering, and preservation. We have obtained various data transformation results, the two most important of which are as follows: standard input, a standard NLI input containing prerequisites and hypotheses as input features ($X$); and null input, an empty string ($\emptyset$) that does not provide any information and is essential for calculating the model prior explicit predictive ability in the absence of explicit evidence.

**PVI Calculator:** This module is responsible for calculating the $\mathcal{V}$-entropy and PVI of the dataset to quantify the amount of information in each data instance. The computational process for PVI and $\mathcal{V}$-information is shown in Algorithm 1. PVI measures the gain in the model predictive confidence after receiving the standard input $x$ compared to receiving null input. According to the definition of Xu et al. [25], let $X$ and $Y$ represent random variables with instance space $\mathcal{X}$ and $\mathcal{Y}$, respectively. Let $\emptyset$ represent an empty input that does not provide information about $Y$.

Let $\Omega = \{f : \mathcal{X} \cup \{\emptyset\} \to \mathcal{P}(\mathcal{Y})\}$. A predictive family is a set of predictive models that the agent is allowed to use; we say that $\mathcal{V} \subseteq \Omega$ is a predictive family if it satisfies

$$\forall f \in \mathcal{V}, \forall P \in range(f), \exists f^{'} \in \mathcal{V}, \forall x \in \mathcal{X}, f'[x] = P, f'[\emptyset] = P, \tag{1}$$

Given the predictive family, the predicted $\mathcal{V}$-entropy is

$$H_{\mathcal{V}}(Y) = \overset{inf}{\underset{f \in \mathcal{V}}{}} \mathbb{E}[-log_2 f[\emptyset](Y)], \tag{2}$$

and the conditional $\mathcal{V}$-entropy is

$$H_{\mathcal{V}}(Y|X) = \overset{inf}{\underset{f \in \mathcal{V}}{}} \mathbb{E}[-log_2 f[X](Y)], \tag{3}$$

where $log_2$ is used to measure the entropy of information bits.

$$I_\mathcal{V}(X \to Y) = H_\mathcal{V}(Y) - H_\mathcal{V}(Y|X), \tag{4}$$

Shannon's mutual information $I(X; Y)$ serves as a direct analog to accessible information [26]. It quantifies the amount of information obtained about one random variable by observing another. $I(X; Y)$ measures the reduction in uncertainty about $Y$ given $X$, assuming an observer with unbounded computational power. Information can be "accessible" in a statistical sense but not practically extractable or actionable by a computationally constrained agent. Useful information is the subset of accessible information that a specific, computationally constrained observer or model $\mathcal{V}$ can actually extract, process, and leverage to perform a given task. It is the information that is actionable and predictive for a defined computational agent. PVI is built on the theory of $\mathcal{V}$-information [25], with the $\mathcal{V}$-information $I_\mathcal{V}(X \to Y)$ in Formula (4) being the difference between the $\mathcal{V}$-entropy $H_\mathcal{V}(Y)$ and the conditional $\mathcal{V}$-entropy $H_\mathcal{V}(Y|X)$. $I_\mathcal{V}(X \to Y)$ reflects how much the model $\mathcal{V}$ can reduce its uncertainty about $Y$ given $X$. The $\mathcal{V}$-entropy measures the uncertainty of the model in predicting labels without input, while the conditional $\mathcal{V}$-entropy measures the uncertainty with input $X$. High PVI indicates that the instance is "simpler" for the model, as input $x$ provides more effective information for the correct prediction of $y$. Therefore, an instance $(x, y)$ is defined as a simple instance if $PVI(x \to y) > \tau$, where $\tau$ is a threshold determined based on the specific task and model performance. Let $D$ be a dataset; a subset $D_{easy} \subseteq D$ containing many simple instances is considered redundant if it satisfies the following condition:

$$\lim_{|D'_{easy}| \to |D_{easy}|} Performance(\mathcal{V}, D) - Performance(\mathcal{V}, D - D'_{easy}) \leq \epsilon_{perf}, \tag{5}$$

After the model $\mathcal{V}$ is trained on $D$ and $D - D'_{easy}$ separately, the performance gap $\nabla Performance$ (e.g., accuracy, F1 score on the validation or test set) resulting from removing more simple instances $D'_{easy} \subseteq D_{easy}$ from the training data tends to an acceptable minimum value $\epsilon_{perf}$. The acceptable removal rate for $D'_{easy}$ adapts to different tasks and datasets, a principle incorporated into the reduction ratios $r$ of Algorithms 2 and 3. Consequently, changes to $D'_{easy}$ directly affect the model's performance; see Section 3.2 for a detailed analysis.

According to the definition of Kawin Ethayarajh [18], the calculation formula for the PVI of an instance $(x, y)$ is as follows:

$$PVI(x \to y) = -log_2 g[\emptyset](y) + log_2 g'[x](y), \tag{6}$$

where $g'$ and $g$ are the models selected from the predictive family $\mathcal{V}$; for example, they can be BERT-family models finetuned under both standard input $(x)$ and null input $(\emptyset)$. Actually, according to Xu et al.'s definition [25], the $\mathcal{V}$ is a mapping set from the input space to the output probability distribution; i.e., $\mathcal{V} \subseteq \{f : X \to P(Y)\}$, where $P(Y)$ is the probability distribution set on the output space $Y$. $g$ usually corresponds to the best predictive function that the model can achieve in the $\mathcal{V}$ family given the input $X$. $g'$ represents the best possible predictor for output $Y$ without a specific input $X$. $g'[x](y)$ is the logarithm of the probability that the model predicts $y$ as the correct label after seeing standard input $x$. $g[\emptyset](y)$ is the logarithm of the probability that the model predicts $y$ as the correct label seeing the null input.

The PVI calculator module receives standard input and null input datasets generated by the data transformation module, along with a pre-trained text classification model (e.g., Chinese-BERT-wwm [27], BERT-base-Chinese [28], and Chinese-MacBERT [29]) and the tokenizer. For each instance, the module calculates its log-likelihood $H_{yb}$ corresponding to $log_2 g'[x](y)$ and $H_{yx}$ corresponding to $log_2 g[\emptyset](y)$ in Formula (6) for standard and null inputs. Finally, the module outputs a series of quantified metrics for each instance, including the PVI, and sorts the instances in the dataset by PVI.

**Reduction Approach:** After obtaining the PVI for all training instances, this module is responsible for conducting the data reduction strategies and evaluating their effectiveness. We have designed two reduction methods, implemented, respectively, by Algorithm 2 and Algorithm 3 (see Section 2.2 for details). Static reduction (Algorithm 2) is a method that aims to evaluate the value of difficult instances. It filters out the subset of instances with low PVI based on a reduction ratio $r$, trains the Chinese-BERT-wwm model from scratch using the subset, and finally evaluates its accuracy on the test set. Progressive learning (Algorithm 3) is a method that adopts a strategy similar to curriculum learning [30], designed to improve training efficiency. It first allows the model to learn from simple instances with high PVI, and then gradually introduces more difficult instances. Finally, it evaluates the accuracy, precision, recall, and F1 score on the test set.

We utilize cross-entropy loss [31] as the optimization objective for model training. Specifically, for a training batch containing $N$ instances, the loss function $J(\theta)$ is defined as follows:

$$J(\theta) = L_{batch} = -\frac{1}{N}\sum_{i=1}^{N}\sum_{c=1}^{C} y_{ic}\log\left(\hat{y}_{ic}\right), \tag{7}$$

where $\theta$ represents the trainable parameters of the model. $N$ is the number of instances in the current training batch. $C$ is the total number of categories, and in an NLI task, $C$ equals 3. $\hat{y}_{ic}$ is the probability that the model predicts that the $i$th instance belongs to category $C$.

The training objective of the model is to find a set of parameters $\theta$ that minimizes the value of the loss function $J(\theta)$:

$$\arg\min_{\theta} J(\theta) \tag{8}$$

### 2.2. Algorithm

Algorithm 1 is the computational process for PVI and $\mathcal{V}$-information. The PVI measures the instance difficulty by comparing the change in confidence of the model predictions for an instance. An instance with high PVI is typically easy to predict, whereas low PVI indicates that the model finds inference for the instance more challenging.

Algorithm 1 calculates the total amount of information provided by input features to the prediction of the target label from the model's perspective. By comparing the predictive capabilities of $g$ and $g'$, the algorithm can analyze the gain of input feature $x_i$ in correctly predicting the model's label. $g$ represents the baseline predictive capability of the model without input features, while $g'$ represents the model's predictive capability conditioned on input features.

---

**Algorithm 1:** PVI Calculator After finetuning on a dataset of size $n$, the $\mathcal{V}$-information and PVI can be calculated in $O(n)$ time

**Input:** training data $D_{\text{train}} = \{(\text{input } x_i, \text{gold label } y_i)\}_{i=1}^{m}$, held-out data $D_{\text{test}} = \{(\text{input } x_i, \text{gold label } y_i)\}_{i=1}^{n}$, model $\mathcal{V}$

**do**

    $g' \leftarrow$ Finetune $\mathcal{V}$ on $D_{train}$

    $\emptyset \leftarrow$ empty string (null input)

    $g \leftarrow$ Finetune $\mathcal{V}$ on $\{(\emptyset, y_i) | (x_i, y_i) \in D_{train}\}$

    $H_{\mathcal{V}}(Y),\ H_{\mathcal{V}}(Y|X) \leftarrow 0, 0$

    **for** $(x_i, y_i) \in D_{test}$ **do**

        $H_{\mathcal{V}}(Y) \leftarrow H_{\mathcal{V}}(Y) - \frac{1}{n}log_2 g[\emptyset](y_i)$

        $H_{\mathcal{V}}(Y|X) \leftarrow H_{\mathcal{V}}(Y|X) - \frac{1}{n}log_2 g'[x_i](y_i)$

        $PVI(x_i \rightarrow y_i) \leftarrow -log_2 g[\emptyset](y_i) + log_2 g'[x_i](y_i)$

    **end for**

    $\hat{I}_{\mathcal{V}}(X \rightarrow Y) = \frac{1}{n}\sum_i PVI(x_i \rightarrow y_i) = H_{\mathcal{V}}(Y) - H_{\mathcal{V}}(Y|X)$

We quantify the amount of information that different datasets provide to the model. Figure 2 illustrates the results of different Chinese datasets providing varying amounts of information to the same model, Chinese-BERT-wwm. According to the distribution of dataset difficulty, the OCNLI dataset contains more information usable by Chinese-BERT-wwm compared to the CMNLI and CINLI datasets, making the computation based on Chinese-BERT-wwm easier.

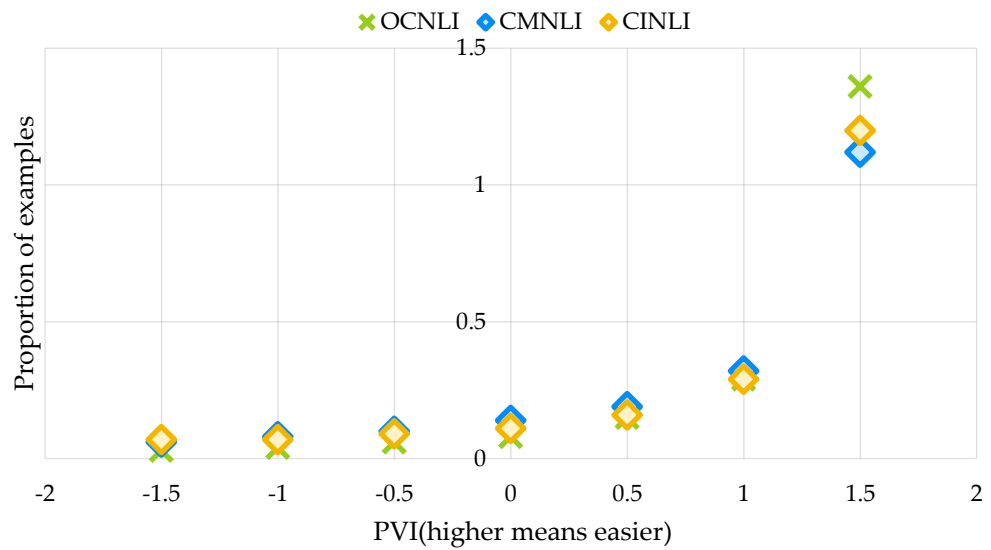

**Figure 2.** The distribution of instance difficulty (PVI) in the held-out sets for each dataset.

Algorithm 2 aims to investigate the relationship between the difficulty of training instances and the model performance through a static data reduction method. Its objective is to evaluate the necessity or redundancy of the simple instances during the model training process and to validate a hypothesis: training the model exclusively with the instances deemed difficult by the model can effectively enhance its generalization ability. The algorithm employs a static strategy, meaning that each experiment uses a fixed, preselected data subset based on a specific difficulty threshold to train a completely new model from scratch. Algorithm 2 first performs PVI computation and difficulty sorting on the entire dataset, using a model finetuned on the full training set. With this model, it calculates the corresponding PVI for each instance $(x_i, y_i)$ in $D_{train}$. After computation, the entire training set is sorted in descending order based on PVI, so that the simple instances with high PVI are at the head of the list, while the difficult instances with low PVI are at the tail. $m$ represents the total number of instances in the training dataset $D_{train}$. Algorithm 2 is a cyclic process that iterates through a series of reduction ratios $r$ (from 0.1 to 0.9). In each iteration, the subset size to be retained is calculated based on the reduction ratio $r$. As $r$ increases, $subset_{size}$ decreases accordingly, meaning that the selected subset contains fewer instances but high average difficulty. To maintain the original batch processing order during training, the selected subset is reordered based on its original indices to obtain the training subset. For each difficult data subset $D_{subset}$ generated through different reduction ratios $r$, the algorithm initializes a completely new, untrained model $model_r$, which is finetuned exclusively using the corresponding $D_{subset}$. After training, the accuracy of $model_r$ is evaluated on the held-out test set $D_{test}$, and the performance under this reduction ratio is recorded (see Section 3.2.1 for details).

---

**Algorithm 2:** Static reduction PVI-based static data reduction for accuracy analysis

**Input:** original training data $D_{\text{train-original}}$, held-out data $D_{test}$, model $\mathcal{V}$, pre-prepared imbalanced training data $D_{\text{train-imbalanced}}$, pre-prepared noisy training data (noisy level = 0.1) $D_{\text{train-noisy}}$

**do**

    **for each** $D_{train}$ **in** [original, imbalanced, noisy] **do**

        $g' \leftarrow$ Finetune $\mathcal{V}$ on $D_{\text{train}}$

        Calculate $PVI(x_i \rightarrow y_i)$ **for** all $(x_i, y_i) \in D_{train}$

        $D_{train}$ sorted $\leftarrow$ Sort $D_{\text{train}}$ instances by PVI in descending order

        **for** $r$ **in** [0.1, 0.2, ... ,0.9] **do**

            $subset_{size} \leftarrow m^1 * (1 - r)$

            $D_{subset}$ sorted $\leftarrow$ Select the last $subset_{size}$ instances from $D_{\text{train}}$ sorted

        $D_{subset} \leftarrow$ reorder $D_{subset}$ sorted by original_idx_i

            $model_r \leftarrow$ Initialize a new model

            Finetune $model_r$ on $D_{subset}$

            Evaluate $model_r$ on $D_{test}$ and record $ACC$ for reduction ratio $r$

        **end for**

    **end for**

**end do**

---

Algorithm 3 aims to explore a progressive learning approach, which is inspired by the concept of curriculum learning. The objective of this algorithm is to validate the hypothesis that by carefully arranging the order of the training instances, easy first, then hard, it can optimize the finetuning process of large language models [32] (https://huggingface.co/Qwen/Qwen3-0.6B (accessed on 15 June 2025), Qwen3-0.6B [33]), thereby achieving faster convergence and better generalization.

Unlike Algorithm 2, which analyzes model performance by statically removing data, Algorithm 3 focuses on dynamically and incrementally feeding data to the model. It first utilizes PVI to rank the entire training set in terms of difficulty, then starts training from the simple instances, and gradually expands the training set to include more difficult instances. Algorithm 3 organizes the instances in an ordered manner from simplest (highest PVI) to most difficult (lowest PVI) from the model's perspective. After each progressive training stage is completed, the model $(model_r)$ trained on the data subset of that stage is evaluated on the held-out test set $D_{test}$. For detailed performance analysis, the evaluation metrics include accuracy, precision, recall, and F1 score. By recording and comparing these metrics at different stages, it becomes clear how model performance evolves as the difficulty and quantity of training data increase (see Section 3.2.2 for details).

---

**Algorithm 3:** Progressive learning PVI-based data reduction and progressive learning for detailed performance evaluation

**Input:** original training data $D_{\text{train-original}}$, held-out data $D_{test}$, model $\mathcal{V}$, pre-prepared imbalanced training data $D_{\text{train-imbalanced}}$, pre-prepared noisy training data (noisy level = 0.1) $D_{\text{train-noisy}}$

**do**

    **for each** $D_{train}$ **in** [original, imbalanced, noisy] **do**

        $g' \leftarrow$ Finetune $\mathcal{V}$ on $D_{\text{train}}$

        Calculate $PVI(x_i \rightarrow y_i)$ **for** all $(x_i, y_i) \in D_{train}$

        $D_{train}$ sorted $\leftarrow$ Sort $D_{\text{train}}$ instances by PVI in descending order

        **for** $r$ **in** [0,0.1,0.2,0.3] **do**

            $subset_{size} \leftarrow m * (1 - r)$

            $D_{subset} \leftarrow$ Select the last $subset_{size}$ instances from $D_{train}$ sorted

            $model_r \leftarrow$ Initialize a new model

            Finetune $model_r$ on $D_{subset}$

Evaluate $model_r$ on $D_{test}$ and record Accuracy, Precision, Recall, F1 for reduction ratio $r$

**end for**

**end for**

**end do**

## 3. Experiments and Results

### 3.1. Experimental Setup

**Dataset:** We utilized three Chinese natural language inference datasets: OCNLI, CMNLI, and CINLI. All datasets contain premise–hypothesis pairs as input features and are annotated with entailment, contradiction, or neutral labels. OCNLI [34] (Original Chinese Natural Language Inference dataset) contains approximately 56,000 premise–hypothesis pairs, entirely based on original Chinese materials. CMNLI [35] (Chinese Multi-Genre Natural Language Inference dataset) integrates Chinese data from XNLI [36] and MultiNLI [37], covering various genres such as news and fiction, used to evaluate cross-domain NLI capabilities. CINLI (Chinese Idioms Natural Language Inference Dataset) focuses on NLI tasks involving Chinese idioms and colloquialisms, containing 91,247 manually annotated idiom pairs, designed to assess models' understanding of subtle semantic differences in Chinese. Before the experiments, we preprocessed the datasets, removing corrupted or incorrectly formatted pairs. The statistical information of the datasets used in the experiments is shown in Table 1, which summarizes the scale and label category statistics for each dataset.

**Table 1.** Category statistics of dataset usage quantity.

| Dataset | Set | Total | Entailment [1] | Neutral | Contradiction |
|---------|-----|-------|----------------|---------|---------------|
| OCNLI | training | 40,340 | 13,464 (33.4%) | 13,734 (34.0%) | 13,142 (32.6%) |
| | testing | 10,097 | 3315 (32.8%) | 3448 (34.1%) | 3334 (33.0%) |
| CMNLI | training | 391,783 | 130,612 (33.3%) | 130,555 (33.3%) | 130,616 (33.3%) |
| | testing | 12,241 | 4277 (32.9%) | 3926 (32.0%) | 4038 (32.9%) |
| CINLI * | training | 80,124 | 26,112 (32.5%) | 26,886 (33.5%) | 27,126 (33.8%) |
| | testing | 26,708 | 8634 (32.3%) | 9022 (33.7%) | 9052 (33.8%) |

\* CINLI is an open-source dataset maintained by individuals, which can be accessed through the GitHub repository (https://github.com/liucongg/NLPDataSet (accessed on 16 June 2025)). [1] The goal of the NLI task is to determine the logical relationship between hypothesis and premise, including three categories of relationships: entailment, neutral, and contradiction.

**Hyperparameter Setting:** For the Chinese-BERT-wwm model, the maximum sequence length is set to 128 tokens, ensuring both the integrity of model input and the optimization of computational resource utilization. The batch size is set to 32, enabling good parallel processing capabilities on most common hardware configurations. The learning rate is set to $5 \times 10^{-5}$, which is a common starting value for finetuning BERT-series models, balancing the model's convergence speed with final performance. The training period is set to 2 epochs, and a linear learning rate scheduler is selected to effectively manage the dynamic changes in the learning rate. Additionally, the gradient accumulation step is set to 1, with gradient updates performed independently for each batch. To ensure the reproducibility of experimental results, a fixed random seed of 1 is set.

The hyperparameter settings for the Qwen3-0.6B model differ to accommodate its model architecture characteristics. The maximum sequence length is extended to 512 tokens to handle longer context information. The batch size is uniformly set to 8, balancing training efficiency and resource consumption under memory-limited conditions. The learning rate is set to $2 \times 10^{-5}$, accompanied by a weight decay of 0.01, to achieve more stable training convergence and prevent overfitting. The model is also trained for 2 epochs, with evaluation and saving strategies set to execute after each epoch ends, facilitating periodic monitoring of model performance and saving the best checkpoints. The logging step is set to 50, enabling fine-grained tracking of the training process. To improve training efficiency and reduce GPU memory usage, mixed-precision training (fp16 = True) is enabled.

### 3.2. Result Analysis

#### 3.2.1. Static Reduction

According to the PVI theory [18], we conducted difficulty analysis and static reduction experiments on the Chinese NLI datasets OCNLI, CMNLI, and CINLI. The theory indicates that high-PVI instances suggest that the model can easily extract information strongly associated with the label $y$ from the input $x$. These instances may contain annotation artifacts (such as high-frequency words, fixed patterns) or shallow patterns, leading the model to achieve high accuracy through "shortcut learning" rather than deep semantic inference. Therefore, removing such instances can encourage the model to learn from low-PVI instances that require more complex inference, thereby enhancing generalization ability and reducing reliance on artifacts.

In the experiment, the Chinese-BERT-wwm model was used to calculate the PVI of the training set, and high-PVI instances were reduced in descending order of PVI by 10%, 20%, ..., 90%, respectively, to construct training subsets with 90%, 80%, ..., 10% of the original size. A series of experiments were conducted, we focused on analyzing the accuracy changes in the classification model at different reduction ratios in Table 2, and Tables 3–5 record the accuracy results on different datasets with different models, where SIM represents the **s**tandard **i**nput **m**odel, EIM represents the **e**mpty **i**nput **m**odel, and CM represents the **c**lassification **m**odel. As the reduction ratio of high-PVI instances increases, the accuracy of the classification models on the three datasets generally shows a declining trend, but the rate and extent of the decline vary across datasets, revealing the moderating effect of different task types on data redundancy. Figure 3 shows this trend.

**Table 2.** Accuracy (%) of CM comparison in each dataset.

| Dataset | $r$ = 0 | 0.1 | 0.2 | 0.3 | 0.4 | 0.5 | 0.6 | 0.7 | 0.8 | 0.9 |
|---|---|---|---|---|---|---|---|---|---|---|
| OCNLI | 69.59 | 68.85 | 66.20 | 62.60 | 54.24 | 49.37 | 41.28 | 34.42 | 26.80 | 22.97 |
| CMNLI | 79.99 | 79.94 | 79.23 | 79.03 | 76.94 | 61.99 | 34.94 | 37.30 | 23.29 | 17.27 |
| CINLI | 91.14 | 91.31 | 90.76 | 89.04 | 87.13 | 77.70 | 79.05 | 75.53 | 48.70 | 48.70 |

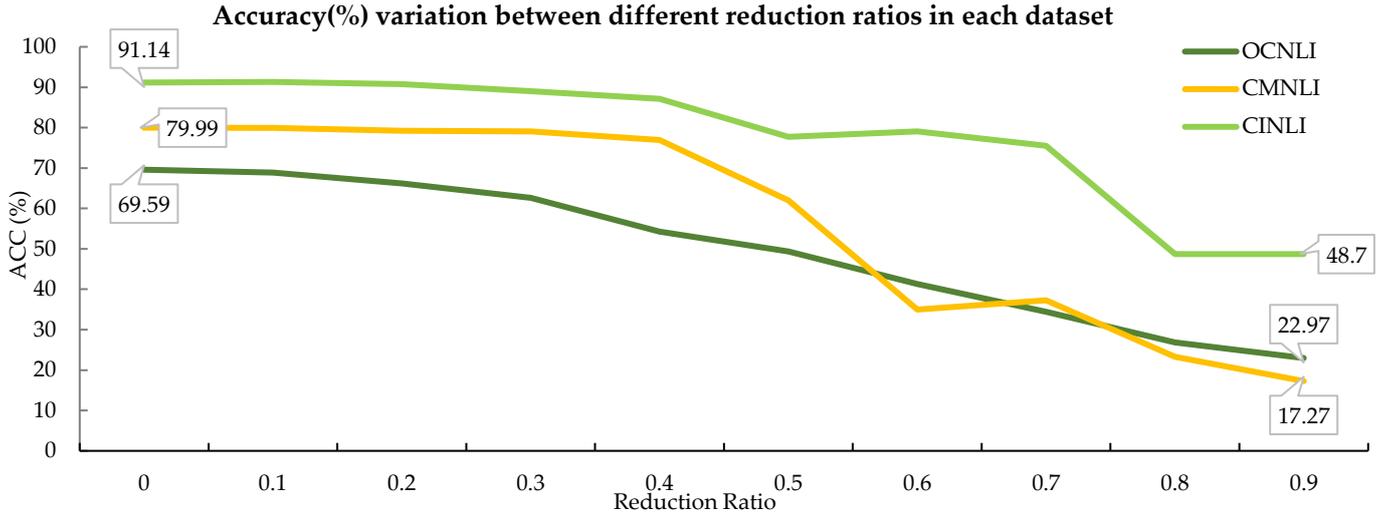

**Figure 3.** Accuracy varies with the reduction ratio. The x-axis represents the reduction ratio (the proportion of data removed), ranging from 0 to 0.9 (or 0% to 90%). The y-axis shows the accuracy (%), indicating the model's performance on the test set. The legend includes three datasets: OCNLI, CMNLI, and CINLI. The figure visually demonstrates that initial data reductions have limited impact on accuracy, especially in the early stages. Our finding implies that training a classifier on the subset with 10%–30% of the original data removed could preserve most of the accuracy while significantly improving training efficiency when combined with an effective data reduction strategy.

**OCNLI:** As the easy instances are reduced, the accuracy of the model on the test set gradually decreases from 69.59% of the full training set to 22.97% (marked in <span style="color:red">red</span> font in Table 3), with performance loss increasing linearly with the proportion of training set reduction. When removing 10%–20% of high-PVI instances, the accuracy of the model decreases slightly (69.59% →68.85%→66.2%), indicating limited dependence of model performance on a small number of high-PVI instances. At this stage, the reduced dataset can save training resources while maintaining model performance within an acceptable range. After reducing 10% of the data, training time decreases, but accuracy drops by only 0.74%, meeting the practical application requirements for balancing efficiency and effectiveness. When 50% of the high-PVI instances are removed, the model accuracy drops to 49.37% (marked in <span style="color:blue">blue</span> font in Table 3), representing a decrease of 19.48% compared to removing 10% of the instances. This indicates that high-PVI instances still contain key generalizable information for the task, and excessive removal can disrupt the model's ability to learn fundamental semantic patterns. The reason might be that not all high-PVI instances correspond to artifacts; some high PVI may arise from genuine strong correlations between input and labels (e.g., the logical relationship of "raining→wet ground" with "entailment" labels), and removing these instances would lead to information loss. Additionally, low-PVI instances contain complex inference patterns but may also include labeling noise or semantic ambiguity. Excessive removal of high-PVI instances alters the data distribution, directly increasing task difficulty beyond the model's processing capacity, resulting in performance collapse.

**Table 3.** Accuracy (%) comparison between different reduction ratios ($r$ from 0 to 0.9) in OCNLI.

| OCNLI | base | 0.1 | 0.2 | 0.3 | 0.4 | 0.5 | 0.6 | 0.7 | 0.8 | 0.9 |
|-------|------|------|------|------|------|------|------|------|------|------|
| SIM* | 89.29 | 83.45 | 83.08 | 78.20 | 66.56 | 64.04 | 57.70 | 57.52 | 63.68 | 72.11 |
| EIM* | 34.12 | 37.09 | 42.57 | 46.85 | 45.94 | 45.53 | 31.98 | 44.95 | 45.27 | 45.94 |
| CM* | <span style="color:red">69.59</span> | 68.85 | 66.20 | 62.60 | 54.24 | <span style="color:blue">49.37</span> | 41.28 | 34.42 | 26.80 | <span style="color:red">22.97</span> |

* SIM: **s**tandard **i**nput **m**odel. EIM: **e**mpty **i**nput **m**odel. CM: **c**lassification **m**odel.

Experiments demonstrate that high-PVI instances are irreplaceable for training the OCNLI model when the reduction ratio $r \geq 0.1$, for the following reasons:

1. Loss of fundamental features: High-PVI instances typically contain strong association patterns between labels and inputs (e.g., the mapping of negation words like "不 (No)" to contradiction-class labels), which serve as the foundation for the model to learn basic inference rules. Removing these patterns makes it difficult for the model to learn basic inference rules.

2. Increased exposure to noise: Potential labeling errors or semantic ambiguity in low-PVI instances (e.g., ambiguous instances labeled as "neutral") are amplified during training, disrupting the model's optimization direction [38]. The removal of high-PVI instances disrupts the stable state of the original data distribution, where the noise dominates the training data, leading the model to converge to local optima. This result validates the core tenet of $\mathcal{V}$-information theory: the difficulty of a dataset is a dynamic function of model capability and data distribution. The removal of high-PVI instances alters the data distribution, thereby changing the task difficulty.

OCNLI is a low-structured task, necessitating the retention of more high-PVI instances to maintain basic inference capabilities. When reducing the data, attention must be paid to the safe reduction ratio $r$ of low-proportion deletion. Removing 10%–20% of high-PVI instances results in only a slight decrease in accuracy on the test set (2–3% drop), making a reduction ratio of around 10% more recommended. The removal of a small number of high-PVI instances can eliminate some redundant artifacts (e.g., overly obvious syntactic templates), prompting the model to learn more generalizable features. However, the reduction ratio must be strictly limited (<20%), and a conservative reduction strategy should be adopted. Beyond 20%, the combined effect of fundamental feature loss and increased noise exposure would accelerate performance decline.

**CMNLI:** Without considering the balance of the dataset, as the reduction ratio increases, the accuracy of the model on the test set gradually decreases, from 79.99% when using the complete training set to 17.27% after removing 90% of easy instances (marked in red font in Table 4), which indicates that many easy instances being removed negatively impacts model performance. Among these, when 10% of high-PVI instances are removed, the accuracy is 79.94%, when 20% are removed, it is 79.23%, and when 30% are removed, it is 79.03% (marked in blue font in Table 4). This is similar to the experimental results on the OCNLI dataset, suggesting that the model's performance has limited dependence on a small number of high-PVI instances. At this point, trimming the dataset can save training resources to some extent while maintaining model performance within an acceptable range. However, when more than 50% of the high-PVI instances are removed, the accuracy drops significantly, such that when 50% are removed, the accuracy is 0.6199 (marked in green font in Table 4), which is 17.95% lower than when 10% are removed. This may be because excessive removal leads to the loss of basic features, making it difficult for the model to effectively learn the semantic patterns, and the noise in low-PVI instances is amplified, affecting the model's optimization direction.

**Table 4.** Accuracy (%) comparison between different reduction ratios ($r$ from 0 to 0.9) in CMNLI.

| CMNLI | Base | 0.1 | 0.2 | 0.3 | 0.4 | 0.5 | 0.6 | 0.7 | 0.8 | 0.9 |
|---|---|---|---|---|---|---|---|---|---|---|
| SIM | 88.58 | 87.06 | 84.66 | 82.01 | 74.52 | 52.75 | 48.00 | 40.74 | 51.65 | 64.70 |
| EIM | 33.34 | 36.36 | 36.43 | 36.22 | 36.93 | 37.97 | 38.90 | 39.93 | 40.77 | 40.71 |
| CM | 79.99 | 79.94 | 79.23 | 79.03 | 76.94 | 61.99 | 34.94 | 37.30 | 23.29 | 17.27 |

**CINLI:** Through the use of the same static reduction method, the top 10%, 20%, ..., 90% of high-PVI instances were removed in descending order of PVI to construct training subsets. The experimental results show that even after removing 40% of the high-PVI instances, the model's accuracy remained at a high level of 87.13% (marked in <span style="color:red">red</span> font in Table 5). This phenomenon contrasts significantly with the OCNLI experimental results, indicating a much slower performance degradation compared to OCNLI, revealing the regulatory effect of task types on data redundancy.

**Table 5.** Accuracy (%) comparison between different reduction ratios ($r$ from 0 to 0.9) in CINLI.

| CINLI | Base | 0.1 | 0.2 | 0.3 | 0.4 | 0.5 | 0.6 | 0.7 | 0.8 | 0.9 |
|---|---|---|---|---|---|---|---|---|---|---|
| SIM | 97.32 | 97.03 | 96.31 | 95.28 | 92.32 | 88.17 | 86.07 | 80.07 | 59.93 | 59.93 |
| EIM | 29.07 | 37.61 | 42.32 | 47.93 | 55.92 | 64.82 | 56.47 | 46.04 | 28.53 | 36.34 |
| CM | 91.14 | 91.31 | 90.76 | 89.04 | <span style="color:red">87.13</span> | 77.70 | 79.05 | 75.53 | 48.70 | 48.70 |

The stability of CINLI stems from its intrinsic characteristics:

1. Structured semantics: The fixed meaning of idioms allows the model to perform generalization inference using a small number of keywords (e.g., "剑 (sword)" in "刻舟求剑", which literally means "to carve a mark on a boat to find a lost sword"; or "蛇 (snake)" and "足 (foot)" in "画蛇添足", which means "to draw a snake and add feet to it"), reducing reliance on data volume and eliminating the need to learn complex contextual correlations. This differs from the causal chain inference in OCNLI, where complex logical inference also requires more task-specific parameter updates. Additionally, the semantic boundaries of idioms are clear, resulting in higher compactness of data distribution and a more concentrated PVI distribution of training instances (low redundancy in high-PVI instances). Even after removal, the remaining instances still cover core semantic patterns.

2. Pre-training compensation: BERT-wwm has encoded the general semantics of idioms [28], thereby reducing sensitivity to training instances. The idiom inference task in CINLI is highly compatible with BERT's masked language modeling objective, both relying on local semantic correlations. Through large-scale corpora, idioms have learned distributed representations, and model finetuning only requires aligning the label space rather than constructing semantic mappings from scratch. Therefore, even after removing some instances, the model can still leverage prior knowledge for generalization inference. This phenomenon aligns with the discussion in the original text on the task–distribution coupling effect: task difficulty is determined by both data distribution attributes (e.g., degree of semantic structuring) and model prior knowledge.

CINLI corresponds to highly structured tasks, with strong feasibility of data reduction, allowing for the prioritized removal of redundant high-PVI instances, saving resources without affecting performance. For such tasks, an aggressive reduction strategy can be adopted, which can reduce approximately 30% of high-PVI instances.

**Class Balance:** In the process of reducing the dataset, we discovered that as more easy instances were reduced, more class imbalances were introduced in the remaining training subset. Therefore, we artificially controlled for proportional reduction in each category and explored the impact of class balance on model training. The experimental results (see Appendix A) indicate that after applying balanced reduction to the dataset to balance the class distribution, the issue of distribution bias caused by the removal of high-PVI instances was mitigated to some extent. Under this balanced constraint, the accuracy of the trained empty input model (EIM) consistently remained close to the random probability of a three-class classification (33%), which aligns with our assumption about balanced reduction. This suggests that the balanced constraint effectively weakens the impact of label distribution bias but does not alter the information-theoretic nature of the empty model. The limited utilization of input information by the empty model and the stability of its performance further highlight the capability of standard input models in effectively utilizing input information for prediction. Simultaneously, this also indirectly confirms that the performance decline of the standard input model after the removal of high-PVI instances is not due to the model itself becoming completely ineffective, but rather because it loses the effective utilization of key input information.

**Noise:** To verify the robustness and generalization ability of the method, we introduced noise into the OCNLI dataset, aiming to simulate more realistic application scenarios. We randomly replaced instances in the OCNLI training set with low-quality text at a replacement ratio of 0.1. These low-quality texts were generated by rewriting original sentences, incorporating features such as synonym substitution, misspellings, punctuation noise, internet slang, meaningless phrase insertion, and sentence splitting and merging. As depicted in Figure 4, the model's performance trend on the noisy dataset closely mirrors that on both class-imbalanced and class-balanced datasets. According to the previous conclusion, a reduction of at least 10% of the data, as per PVI, has little impact on the model's performance. It is evident that the performance on datasets with added noise almost remains lower than that on the other two dataset types (imbalanced and balanced).

**Random:** We also included a random baseline as a control, randomly deleting instances from the training data without considering any scores or specific features of the data points. We observed that reducing the easy instances did not lead to a gain in performance over the random baseline. A similar phenomenon appeared in Rabiraj's research [39], which designed a pruning strategy for sexism detection using three influence scores including PVI. As the reduction ratio of simple instances increased, the training subset contained more difficult instances. Consequently, the difficulty of the dataset rose, and the performance gap between the random baseline and other baselines using PVI for reduction gradually widened. We speculated that training solely on a difficult subset could cause the model to over-emphasize learning from edge cases and ambiguous instances, potentially leading to overfitting to these specific hard instances and consequently poor generalization on the broader test set. The random baseline's performance, where some easy instances were inevitably retained due to random deletion, might implicitly benefit from this broader representation. Removing a large number of simple instances could disrupt the difficulty distribution of the dataset. Therefore, we recommend reducing the dataset within a moderate range, removing simpler instances only to the extent that the overall difficulty structure remains largely intact. To further improve upon the baseline performance and the reduction in simple instances, one could incorporate a more sophisticated weighting mechanism for the remaining difficult instances during training. In subsequent work, we will consider combining PVI-based difficulty measurement with diversity measurement (e.g., EL2N [40], VoG [41], TracIn [42]) as an improvement strategy to select the instances to be retained. This would ensure that the model learns not only from challenging instances but also from a diverse set of representative instances across the data spectrum.

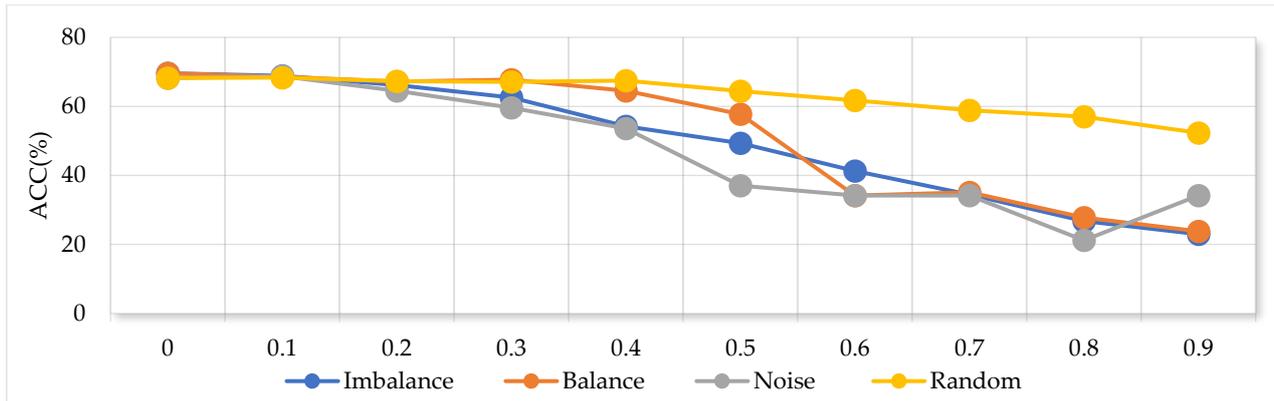

**Figure 4.** Accuracy (%) variation between different ratios in OCNLI.

### 3.2.2. Progressive Learning

In this section, the experiments primarily focus on the OCNLI and CINLI datasets, aiming to investigate the effectiveness of progressive learning strategies. The selection of these two datasets is based on the following considerations: The OCNLI dataset holds significant representativeness in the field of Chinese natural language inference, effectively evaluating the model's baseline performance and generalization capabilities; the CINLI dataset, with its unique text pair construction and inference task design, facilitates an in-depth examination of the model's inference accuracy and stability. In comparison, the CMNLI dataset, with its large instance size and status as a translation-generated dataset, exhibits limitations such as semantic bias and cultural differences, which may introduce confounding factors. Therefore, under constrained experimental resources, prioritizing the OCNLI and CINLI datasets ensures the acquisition of more reference-worthy and persuasive experimental results.

Following Algorithm 3, the training set is sorted based on PVI (from easiest to hardest), and Qwen3-0.6B (available on https://huggingface.co/Qwen/Qwen3-0.6B (accessed on 15 June 2025)) is used as the base model to train. Initially, PVI values are computed for all instances in the training set to establish their difficulty ranking. Then, the training process commences with the simplest instances and gradually incorporates more difficult ones by selecting subsets of the sorted training data. After each progressive training stage on a subset, the trained model is evaluated on a fixed held-out test set, recording accuracy, precision, recall, and F1 score to assess performance evolution. The experimental results demonstrate that training the dataset sorted by PVI enhances model performance. Since Micro-average is used to calculate recall in multi-class tasks, the three categories in the dataset are relatively evenly distributed, with values close to accuracy.

Table 6 presents the experimental results on the OCNLI dataset. By sorting the training set based on PVI from easiest to hardest, the model's accuracy improves by approximately 0.81% relative to the baseline (69.76% − 68.95 = 0.81%), and the F1 score also rises from a baseline of 69.08% to 69.91%, an increase of about 0.83% (marked in **bold** font in Table 6). This indicates a positive impact of PVI sorting on model performance. Even with a 10% reduction in training data, the model performance remains high, reflecting the effectiveness of the sorting and reduction strategies. Table 7 presents the mean and standard deviation of the model's performance over three runs on the OCNLI dataset. Figure 5 visually compares the model's performance under different processing methods. Comparing the "Base" (green bar) and "Sort" (blue bar) clearly shows that after PVI sorting, the model improves in accuracy, precision, recall, and F1 score. While "Sort & Reducing 10%" (orange bar) performs slightly lower than "Sort" on all metrics, it still maintains a level of precision close to that of "Base," consistent with the data analysis in Table 6, further confirming that even with reduced data volume, the model can still exhibit strong performance.

**Table 6.** OCNLI results under the optimal reduction ratio ($r$=0.1).

| Data Processing | Accuracy | Precision | Recall | F1 |
|---|---|---|---|---|
| Base | **68.95** | 70.22 | 68.95 | **69.08** |
| Sort | **69.76** | 70.49 | 69.76 | **69.91** |
| Sort & Reducing 10% | 68.28 | 70.31 | 68.28 | 68.48 |

**Table 7.** Mean ± standard deviation of multiple runs on the OCNLI dataset.

| Data Processing | Accuracy | Precision | Recall | F1 |
|---|---|---|---|---|
| Base | 69.15 ± 0.29 | 70.19 ± 0.15 | 69.15 ± 0.29 | 69.23 ± 0.31 |
| Sort | 70.32 ± 0.48 | 71.02 ± 0.54 | 70.32 ± 0.49 | 70.45 ± 0.47 |
| Sort & Reducing 10% | 68.69 ± 0.63 | 70.25 ± 0.34 | 68.69 ± 0.63 | 68.88 ± 0.64 |

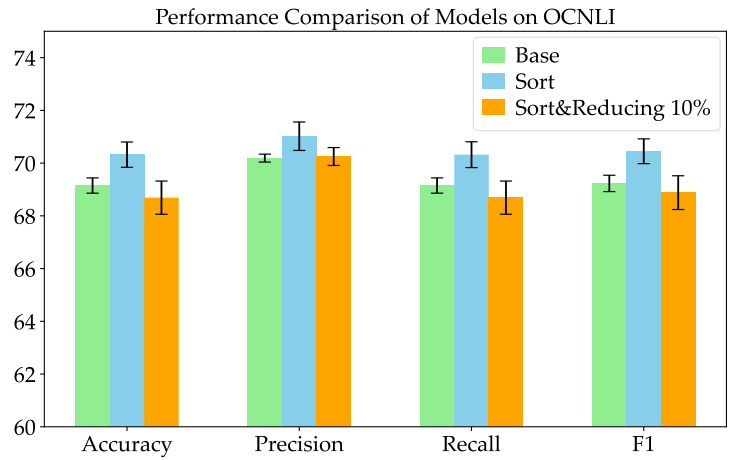

**Figure 5.** Comparison of indicators on the OCNLI dataset.

Table 8 presents the experimental results on the CINLI dataset, showing that the model's performance also improves after data processing. The accuracy increases from the baseline of 91.7852% to 91.8676%. The F1 score rises from the baseline of 91.7861% to 91.8651% (marked in **bold** font in Table 8). At a reduction ratio of $r$=0.3 (i.e., reducing 30% of the data volume), the model still maintains an accuracy of 90.42% and an F1 score of 90.38%, further validating that the progressive learning strategy can effectively reduce the demand for training data while preserving model performance. Figure 6 compares the model's performance on the CINLI dataset under different processing methods. Similarly to the analysis of OCNLI, Figure 6 clearly illustrates the improvements in "Sort" (blue bar) over "Base" (green bar) in all performance metrics, although the magnitude of the improvement is relatively small. It is noteworthy that the performance of "Sort & Reducing 30%" (orange bar) declines in accuracy, precision, recall, and F1 score, but remains above 90%.

**Table 8.** CINLI results under the optimal reduction ratio ($r$ = 0.3).

| Data Processing | Accuracy | Precision | Recall | F1 |
|---|---|---|---|---|
| Base | **91.7852** | 91.7873 | 91.7852 | **91.7861** |
| Sort | **91.8676** | 91.8677 | 91.8676 | **91.8651** |
| Sort & Reducing 30% | 90.4223 | 90.4655 | 90.4223 | 90.3838 |

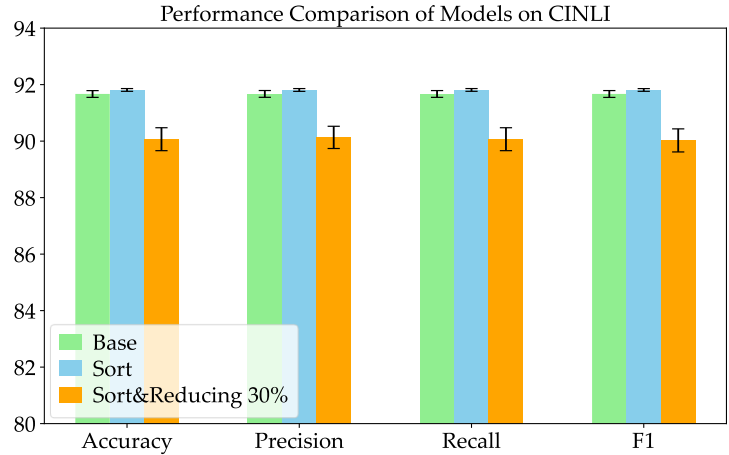

**Figure 6.** Comparison of indicators on the CINLI dataset.

Table 9 presents the mean and standard deviation of the model's performance over three runs on the CINLI dataset.

**Table 9.** Mean ± standard deviation of multiple runs on the CINLI dataset.

| Data Processing | Accuracy | Precision | Recall | F1 |
|---|---|---|---|---|
| Base | 91.67 ± 0.11 | 91.67 ± 0.12 | 91.67 ± 0.11 | 91.66 ± 0.12 |
| Sort | 91.81 ± 0.05 | 91.81 ± 0.05 | 91.81 ± 0.05 | 91.81 ± 0.04 |
| Sort & Reducing 30% | 90.07 ± 0.41 | 90.13 ± 0.39 | 90.07 ± 0.41 | 90.03 ± 0.41 |

We speculate that this progressive learning strategy from easy to difficult (as a form of curriculum learning) enables the model to prioritize learning instances that are information-rich but low in difficulty during the early stages of training, thereby rapidly constructing foundational feature representations and pattern recognition capabilities. Subsequently, the model gradually exposes itself to and learns more complex instances, which helps it progressively master more abstract and fine-grained knowledge. This reasonable distribution of difficulty optimizes the "quality" and utilization efficiency of the training set during the training process, avoiding interference from a large number of difficult or noisy instances in the early stages, thus promoting faster convergence rates and higher final performance. From the perspective of model optimization, a reasonable distribution of difficulty can guide the gradient descent process to converge to better local minima or, at the very least, achieve more robust parameter initialization in the early stages of training, laying a solid foundation for subsequent learning.

### 3.3. Computational Efficiency Analysis

Figure 7 illustrates the empty input model's computation time for each dataset at various reduction ratios, using a single NVIDIA A100 GPU (NVIDIA, Santa Clara, CA, USA). The EIM's runtime decreases as the dataset size is reduced. While data reduction decreases the training time, the initial PVI computation requires some overhead, particularly at lower reduction rates. Where there is an initial cost, we should consider the break-even point and subsequent benefits of training, especially when considering the possibility of larger reductions or multiple training runs on the same reduced dataset. This overhead cost may be acceptable, depending on the overall efficiency gains.

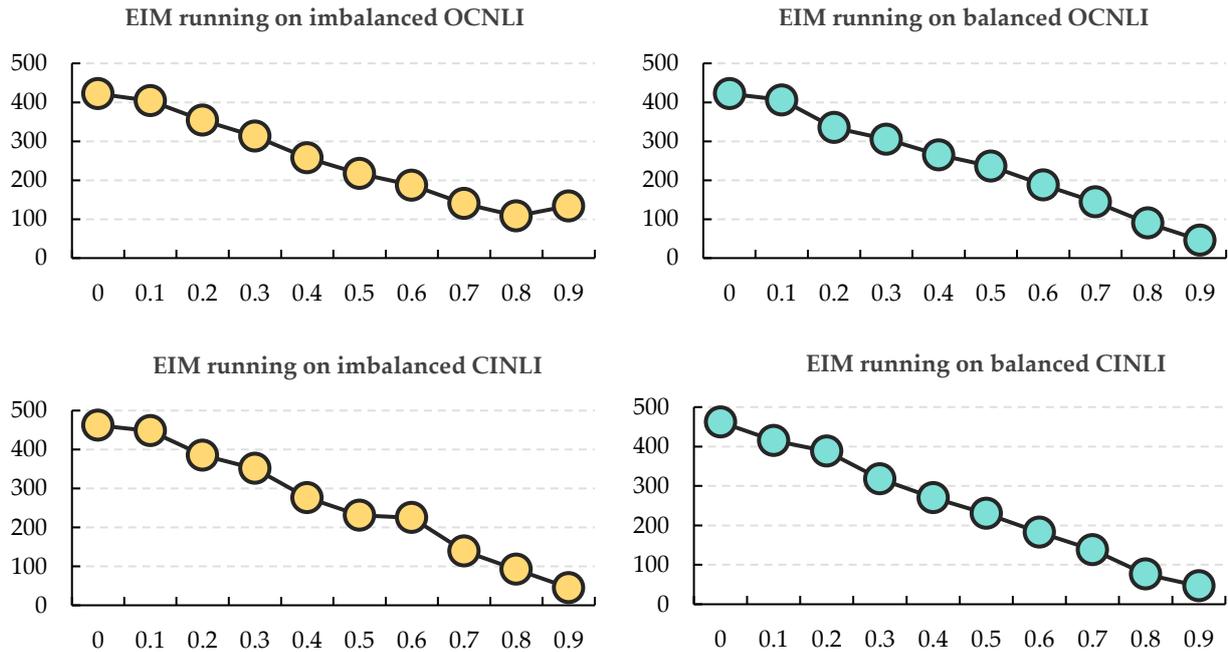

**Figure 7.** Runtime analysis of empty input models during the training process. The x-axis represents the reduction ratio $r$, ranging from 0 to 0.9, indicating the proportion of high-PVI instances removed during data reduction. The y-axis shows the training time (in seconds), reflecting the computational resources required for training under each $r$. The yellow dot indicates the calculation time of running the empty model on the class-unbalanced dataset, and the green dot indicates the calculation time of running the empty model on the class-balanced dataset.

## 4. Discussion

This chapter discusses the reasons why the NLI dataset poses challenges for model construction, which may stem from the inherent characteristics of the dataset and its intrinsic distribution.

In the CMNLI dataset, the distribution of token counts across different inference categories is unbalanced. Figure 8 shows the histogram of hypothesis length distribution in the CMNLI dataset, and combined with the statistical information in Table 10, it can be observed that the statistical features of neutral hypotheses (label 1) are significantly different from other categories, with a median (17.0) and mean (18.35) that are both the highest, and a maximum value reaching 100 tokens. This indicates that the length distribution of neutral hypotheses is right-skewed, reflecting that maintaining semantic neutral status requires more modifiers, such as adding conditional adverbials ("under certain conditions") or hedges ("possibly"), leading to neutral hypotheses containing the longest instances. Therefore, hypothesis length becomes an effective feature, with neutral hypotheses dominating the longer text intervals (e.g., constituting 27.20% in the 16–20-token range, and consistently leading in intervals ≥31 tokens). Contradiction hypotheses (label 2) exhibit the shortest concentration trend, with a median of 15.0. The syntactic characteristic of Chinese, known as "parataxis," allows for the expression of complex logic using fewer tokens, which may have influenced the conciseness of contradiction hypothesis categories.

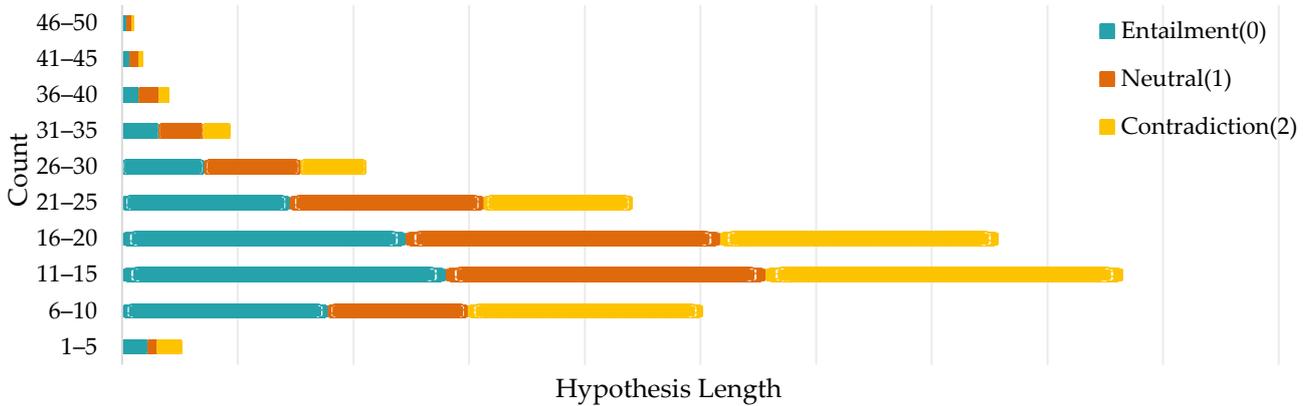

**Figure 8.** Display of hypothesis length distribution for CMNLI.

**Table 10.** Hypothesis information statistics of CMNLI.

| Label | Min | Max | Median | Mean |
|-------|-----|-----|--------|------|
| 0 | 1.0 | 99.0 | 16.0 | 17.1 |
| 1 | 1.0 | 100.0 | 17.0 | 18.3 |
| 2 | 1.0 | 87.0 | 15.0 | 16.1 |

The proportions of assumptions in different length intervals of the CMNLI dataset are presented in Table 11. In the short text interval (1–10 tokens), the proportions of entailment (0.201) and contradiction (0.225) are slightly higher than neutral (0.129). Short texts do not exhibit a clear category advantage, which is related to the characteristics of the Chinese language. In the core distribution interval (11–25 tokens), the neutral hypothesis has the highest proportion (27.20%) in the 16–20-token range, while the contradiction hypothesis forms a peak in the 11–15-token range (30.9%), with this region showing cross-competition among the three types of hypotheses. In the ultra-long text interval (≥31 tokens), the neutral hypothesis maintains a leading proportion, significantly higher than entailment and contradiction.

**Table 11.** The proportions of assumptions in different length intervals of CMNLI.

| Label | ≤10 Tokens | 11–15 Tokens | 16–20 Tokens | 21–25 Tokens | 26–30 Tokens | ≥31 Tokens |
|-------|-----------|--------------|--------------|--------------|--------------|------------|
| 0 | 0.201 | 0.279 | 0.244 | 0.145 | 0.071 | 0.060 |
| 1 | 0.129 | 0.277 | 0.272 | 0.168 | 0.084 | 0.070 |
| 2 | 0.225 | 0.309 | 0.241 | 0.128 | 0.056 | 0.041 |

For OCNLI, as observed from the distribution histogram in Figure 9 and sentence length statistics in Table 12, the distribution of hypothesis lengths in the OCNLI dataset exhibits a clear right skew, with most hypotheses concentrated in shorter length intervals (particularly 5–15 tokens). Compared to the CMNLI dataset, OCNLI's hypotheses are generally shorter, and even the longest instances are significantly shorter (maximum value of 60 tokens). The construction of the OCNLI dataset emphasizes shorter, more direct inference scenarios, and the characteristics of its text sources also contribute to the shorter hypotheses.

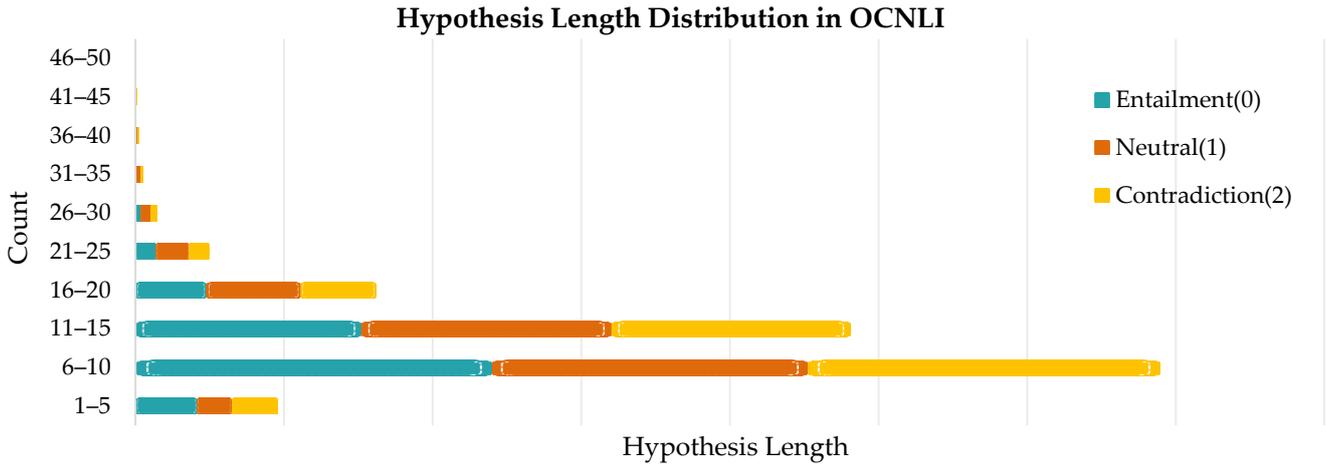

**Figure 9.** Display of hypothesis length distribution for OCNLI.

**Table 12.** Hypothesis information statistics of OCNLI.

| Label | Min | Max | Median | Mean |
|-------|-----|-----|--------|------|
| 0 | 2.0 | 54.0 | 10.0 | 10.7 |
| 1 | 3.0 | 60.0 | 11.0 | 11.9 |
| 2 | 2.0 | 55.0 | 10.0 | 11.0 |

The proportion of different length intervals of assumed content in the OCNLI dataset is presented in Table 13. In the short text interval (1–5 tokens), the proportion of entailment assumptions (0.083) and contradiction assumptions (0.062) is significantly higher than that of neutral assumptions (0.046). This indicates that in the OCNLI dataset, short texts seem to better support entailment and contradiction relationships, which may be related to certain phrases or expressions in Chinese that can directly constitute entailment or contradiction relationships. The core distribution interval (6–15 tokens) is a very concentrated interval, with all three types of assumptions accounting for most instances. The 6–10-token interval is the peak: entailment (0.479), neutral (0.425), and contradiction (0.475) all reach their respective peaks in this interval, with proportions all close to or exceeding 40%. This suggests that the core assumption length in the OCNLI dataset is concentrated between 6 and 10 tokens. In the 11–15-token interval, the proportion of neutral assumptions (0.337) is the highest, slightly exceeding that of contradiction (0.321) and entailment (0.304). This again confirms the trend that neutral assumptions tend to be relatively longer. In the medium–long text interval and the ultra-long text interval, the advantage of neutral assumptions gradually becomes apparent, with proportions consistently leading those of entailment and contradiction assumptions. These results further demonstrate that maintaining neutral status requires longer expressions or neutral inference in more complex contexts.

**Table 13.** The proportions of assumptions in different length intervals of OCNLI.

| Label | 1–5 Token | 6–10 Token | 11–15 Token | 16–20 Token | ≥21 Token |
|-------|-----------|------------|-------------|-------------|-----------|
| 0 | 0.083 | 0.479 | 0.304 | 0.095 | 0.039 |
| 1 | 0.046 | 0.425 | 0.337 | 0.126 | 0.066 |
| 2 | 0.062 | 0.475 | 0.321 | 0.102 | 0.040 |

Furthermore, we have listed the most challenging instances from the OCNLI test set according to Chinese-BERT-wwm, detailed in Appendix B. All three categories—entailment, neutral, and contradiction—are represented in Table B1, with entailment representation appearing slightly excessive. Some instances have actually been incorrectly labeled—for instance, the instance "Premise: 他是去那个南方那个学校嘛 (He is going to that southern school, right?) Hypothesis: 国防动员无需加强 (National defense mobilization does not need to be strengthened)" is labeled as "entailment," although the correct label should be "neutral."

## 5. Conclusions and Future Work

We introduced an effective data reduction strategy based on Pointwise V-Information (PVI) to enhance model training efficiency and performance in data-centric AI. We successfully extended the PVI framework, previously limited to English datasets, to various Chinese NLP tasks and base models, addressing a critical gap in cross-lingual data reduction.

The use of PVI in this article also has certain limitations. A key challenge is that PVI requires the model to produce a full probability distribution over all possible outputs. This can be problematic for tasks where such distributions are not readily available, such as machine translation that relies on beam search. In addition, while PVI enables comparison of different instances with respect to the same attribute (e.g., the amount of token-identity information available for two instances), it does not support direct comparison of different attributes for the same instance.

In future work, we aim to validate the generality of our method in diverse linguistic environments by conducting experiments on German, French, and other multilingual NLI corpora. To this end, we will explore and utilize pre-trained language models suitable for these languages, such as GottBERT (for German) and FlauBERT (for French). This comprehensive evaluation will allow us to gain a deep understanding of the PVI-based strategy's capabilities for data reduction and performance enhancement in cross-lingual scenarios, as well as identify potential challenges across different language characteristics and task contexts.

We acknowledge that the optimal data reduction approach may vary significantly across different data modalities, such as text, images, or tabular data. Therefore, tailoring reduction methods to the unique characteristics of different data types and application domains will be crucial. The information-theoretic basis of PVI indicates its relevance beyond textual data. We anticipate expanding PVI's application to non-text modalities, including visuals and speech. Applying PVI to picture data necessitates delineating the extraction of model-usable information from pixel arrays and the construction of "empty inputs" for visual tasks. Likewise, with audio data, PVI might potentially utilize spectral information or alternative acoustic representations. Extending PVI to these modalities necessitates meticulous evaluation of modality-specific attributes, including the establishment of suitable null inputs or baselines for computing V-information that accurately represent the computational limitations of models functioning with these data types. Comprehending these modality-specific adaptations and constraints will be essential for our future study, aiding in the evaluation of PVI's generalizability as a universal standard for data quality.

Data selection encompasses multiple criteria, such as entropy-based techniques, loss-based data trimming, gradient-based sampling, and methods inspired by active learning, among others. We also plan to move beyond single indicators for data point selection, aiming for more nuanced metrics that capture a data point's value in terms of diversity, informativeness, or representativeness of important subgroups.

Additionally, an exciting direction involves combining data reduction with synthetic data generation. The Phi-series models [43,44] from Microsoft demonstrate that carefully selected and optimized training datasets can substantially decrease model size without compromising performance. Synthetic data, often produced by advanced LLMs, is utilized to augment these datasets, such as by closing performance gaps or delivering specialized skill training. Future systems could identify gaps created by aggressive filtering, especially for rare but important instances, and then use generative models to create synthetic data to fill these specific gaps, ensuring comprehensive coverage.

## Appendix A

**Table A1.** Impact of label distribution bias on the model. The accuracy of the trained empty input model (EIM, marked in bold font in Table A1) consistently remained close to the random probability of a three-class classification (33%).

| Dataset | Model | base | 0.1 | 0.2 | 0.3 | 0.4 | 0.5 | 0.6 | 0.7 | 0.8 | 0.9 |
|---------|-------|------|-----|-----|-----|-----|-----|-----|-----|-----|-----|
| OCNLI | SIM | 89.29 | 86.65 | 80.36 | 79.74 | 71.45 | 57.92 | 34.17 | 59.94 | 63.80 | 70.65 |
| | **EIM** | **34.12** | **33.79** | **33.40** | **33.18** | **33.01** | **32.57** | **32.98** | **34.17** | **33.63** | **34.43** |
| | CM | 69.59 | 68.56 | 67.22 | 67.73 | 64.47 | 57.74 | 34.16 | 35.11 | 27.75 | 23.82 |
| CMNLI | SIM | 88.58 | 87.42 | 84.23 | 81.21 | 74.07 | 42.25 | 33.32 | 34.45 | 33.34 | 61.05 |
| | **EIM** | **33.34** | **33.34** | **33.34** | **33.34** | **33.34** | **32.97** | **33.32** | **33.76** | **33.29** | **33.32** |
| | CM | 79.99 | 79.93 | 78.92 | 78.78 | 76.60 | 47.41 | 32.07 | 33.68 | 34.94 | 19.47 |
| CINLI | SIM | 97.32 | 96.92 | 97.32 | 94.41 | 92.58 | 89.03 | 79.60 | 33.56 | 44.13 | 92.58 |
| | **EIM** | **29.07** | **33.56** | **27.09** | **29.65** | **33.86** | **33.86** | **33.56** | **33.56** | **32.10** | **33.86** |
| | CM | 91.14 | 91.13 | 91.14 | 90.40 | 87.26 | 83.73 | 75.22 | 33.78 | 41.08 | 87.26 |

## Appendix B

Certain instances within Table A2 are assessed to be mislabeled by the authors of this work, and these are visually indicated in red. The labels of the fifth and ninth instances are entailment, which should be changed to neutral. The label of the 11th instance is neutral, which should be changed to entailment.

**Table A2.** Part of the hardest (lowest-PVI) instances in the OCNLI test set for logical inference (label indicates the logical relationship between "premise" and "hypothesis"), according to Chinese-BERT-wwm. Instances in red are assessed to be mislabeled by the authors of this work.

| Num | Premise | Hypothesis | Label | PVI |
|-----|---------|------------|-------|-----|
| 1 | 其中有一个这两天记者采访他还出诊呢 | 记者工作是出诊 | 矛盾 | −8.745 |
| | One of them was interviewed by a journalist and even made house calls these last two days | Journalists make house calls | Contradiction | |
| 2 | 所以对热闹的世界杯充耳不闻 | "我们"没有关注世界杯 | 蕴含 | −7.125 |
| | So I turned a deaf ear to the lively World Cup | "We" did not pay attention to the World Cup | Entailment | |
| 3 | 处理中美关系应着眼于全球,着眼于二十一世纪 | 二十世纪对处理中美关系不重要。 | 中立 | −6.645 |
| | Handling China-US relations should be focused on the global perspective, focused on the 21st century | The 20th century is not important for handling China-US relations | Neutral | |
| 4 | 然而,古巴、也门和其它一些国家一直要求立即取消对伊拉克的制裁 | 其他一些国家包括古巴和也门 | 矛盾 | −6.622 |
| | However, Cuba, Yemen, and some other countries have consistently demanded the immediate lifting of sanctions on Iraq | Other countries include Cuba and Yemen | Contradiction | |
| 5 | 他是去那个南方那个学校嘛 | 国防动员无需加强 | 蕴含 | −6.561 |

| | Is he going to that school in the south | National defense mobilization does not need to be strengthened | Entailment | |
|---|---|---|---|---|
| 6 | 对外承包工程和劳务合作完成营业额近 13 亿美元 | 营业额达到了 13 亿美元 | 矛盾 | −6.545 |
| | Contracted engineering projects and labor cooperation completed nearly $1.3 billion in turnover | Revenue reached $1.3 billion | Contradiction | |
| 7 | 去年 8 月海湾冲突爆发后,日本政府曾向国会提出了一项旨在向海外派兵的联合国和平合作法案 | 日本政府没有独立的立法权 | 蕴含 | −6.333 |
| | Last August after the Gulf conflict broke out, the Japanese government proposed a UN Peace Cooperation Bill to the Diet aimed at deploying troops overseas | The Japanese government does not have independent legislative power | Entailment | |
| 8 | 各项决策都要做到程序依法规范、过程民主公开、结果科学公正 | 没有一条好决策的出台能够脱离依法规范的程序 | 蕴含 | −6.291 |
| | All decisions must be made in accordance with legally standardized procedures, democratic and open processes, and scientifically fair outcomes | No good decision can be implemented without reference to legally standardized procedures | Entailment | |
| 9 | 花篮里的花又白的多红的少,专配银冠似的 | 我对花盆的花颜色的搭配嗤之以鼻 | 蕴含 | −6.260 |
| | The flowers in the basket are mostly white and few red, perfectly matching the silver crown-like appearance | I sneer at the color combination of the flowers in the pots | Entailment | |
| 10 | 合理的投资规模是保持经济稳定和增强发展后劲的重要条件 | 不合理的投资规模制约经济持续向好 | 蕴含 | −5.989 |
| | A reasonable investment scale is an important condition for maintaining economic stability and enhancing development potential | An unreasonable investment scale restricts the economy's sustainable upward trend | Entailment | |
| 11 | 对,出席大会的时候还自我调侃,说这个整个场面,我是个科学家,我不是摇滚明星 | 举办过一场大会 | 中立 | −5.944 |
| | Yes, when attending the conference, I even made a self-deprecating joke, saying that in this entire scene, I'm a scientist, not a rock star | Organized a conference | Neutral | |

## Appendix C

A comprehensive list and description of all symbols employed in this paper can be found in Table A3.

**Table A3.** List of notations and their descriptions.

| Notations | Descriptions |
|---|---|
| $x$ | Sentence |
| $X$ | Random variables taking values in $\mathcal{X}$ |
| $\mathcal{Y}$ | Label |
| $Y$ | Random variables taking values in $\mathcal{Y}$ |
| $(x, y)$ | An instance in the dataset |
| $\mathcal{V}$ | Predictive family |
| $f(\cdot)$ | Predictive function in $\mathcal{V}$ |
| $g$ | The best predictive function for a given input $X$ |
| $g'$ | The best predictive function for output $Y$ without specific input $X$ |
| $\emptyset$ | Empty input |
| $H_V(\cdot)$ | $\mathcal{V}$-entropy |
| $D$ | Complete dataset |
| $D_{easy}$ | Dataset composed of simple instances |
| $D'_{easy}$ | A subset selected from $D_{easy}$ |
| $J(\theta)$ | Loss function |
| $\tau$ | Threshold for defining simple instances |
| $\epsilon$ | Minimum value of performance gap |
| $I(\cdot)$ | Shannon's mutual information |
| $r$ | Reduction ratio |
| $m$ | Total number of instances in the training dataset |
| $\mathcal{P}(\cdot)$ | The set of all probability measures over the Borel algebra |
| $t$ | $X \rightarrow X$, being any function |